\documentclass{article}

\PassOptionsToPackage{numbers, compress}{natbib}

 \usepackage[dblblindworkshop, final]{neurips_2026}
\workshoptitle{New in Machine Learning (NewInML) at NeurIPS 2026}

\usepackage[utf8]{inputenc} 
\usepackage[T1]{fontenc}    
\usepackage{hyperref}       
\usepackage{url}            
\usepackage{booktabs}       
\usepackage{amsfonts}       
\usepackage{nicefrac}       
\usepackage{microtype}      
\usepackage{xcolor}         
\usepackage{amsmath}
\usepackage{graphicx}
\usepackage{IEEEtrantools}

\title{``Close enough?'' An investigation into multiplication errors in LLMs}

\author{%
  William Bolton \\
  School of Engineering Maths \\
  \& Technology \\
  University of Bristol\\
  \texttt{tv24888@bristol.ac.uk} \\
  \And
  Bezawit Tilaye \\
  School of Computer Science \\
  University of Bristol \\
  \And 
  Huw Day \\
  School of Engineering Maths \\
  \& Technology \\
  University of Bristol\\
  \texttt{huw.day@bristol.ac.uk}
}

\begin{document}
\bstctlcite{IEEEexample:BSTcontrol}

\maketitle

\begin{abstract}
    Large language models (LLMs) frequently fail at multi-digit arithmetic, but the underlying mechanisms governing these failures remain poorly understood. While recent work demonstrates that addition errors follow a "noisy quantisation" model localised around carry boundaries, whether this framework generalises to multiplication remains an open question. In this work, we evaluate multiplication capabilities across several small base LLMs on a dataset of 16,991 problems spanning varied operand lengths. We show that multiplication errors do not conform to the addition-style noisy quantisation model. Instead, we uncover a novel error regime in small Qwen models: errors in lower-order columns predominantly preserve parity. Conversely, higher-order digit errors shift toward off-by-one carry failures. Our findings suggest that arithmetic computation in LLMs relies on distinct, multi-periodic internal representations where parity is tracked more robustly than exact digit identity, providing new mechanistic insights into numerical reasoning in transformers.
\end{abstract}

\section{Introduction}
Despite empirical evidence of Large Language Models (LLMs) being useful mathematical tools, helping to solve difficult Olympiad problems \cite{trinh2024alphageometry, chervonyi2025alphageometry2} and more recently resolving open research problems including some Erdős conjectures \cite{tsoukalas2026advancingmathematicsresearchaidriven}, producing a proof for Navier-Stokes and several other problems in mathematics \cite{openai2026navierstokes, openai2026sharingmath}, they are notoriously unreliable at arithmetic \cite{li2025exposingnumeracygapsbenchmark} and unpredictable in how they represent numbers in embedding space \cite{davies2025languagemodelsembednumbers}. This is largely unsurprising given they are probabilistic and designed to produce fluent natural language rather than accurate arithmetic. However, it is concerning given how many people use LLMs as a general oracle of information \cite{brachman2025currentfutureuselarge}. Production LLMs have improved at arithmetic and will often use tooling for larger calculations to guarantee a correct answer \cite{schick2023toolformerlanguagemodelsteach}. Despite this, they are still fallible. 

The way in which they do make errors when not permitted to use tooling is interesting. Arithmetic provides us with a useful playground for understanding how LLMs struggle with precise information in general since it is an easy, verifiable, controlled task. For this reason, lots of work has been done recently to use arithmetic as a way of understanding how LLMs handle precise information. Understanding the mechanisms by which LLMs make errors in arithmetic should allow for the development of models that are better at dealing with precise information.

\paragraph{Contribution:}in this work, we investigate the modes of errors in multiplication over a range of small LLMs and explore whether models for addition failure (such as the `noisy quantisation model' \cite{shapeofaddition}) transfer to multiplication. We then propose future work using mechanistic interpretability techniques to validate our hypotheses.

\section{Related Work}

Rather than simply recalling arithmetic tables from its training data, LLMs are required to do arithmetic. In a 2024 study, the authors found that LLMs use `Fourier features' (periodic, trigonometric functions, sparse in the frequency domain) to represent numbers \cite{zhou2024pretrainedlargelanguagemodels}. Studying a fine-tuned GPT-2-XL alongside in-context-prompted GPT-J and Phi-2, and extending their analysis to the closed source GPT-3.5 and GPT-4, the authors showed that models have learned to represent numbers on an oscillating wave, encoding information about how they relate to each other mathematically. This study also found that different components of the LLM perform different roles. Namely, the MLP layers approximate the magnitude of an answer with low-frequency waves and attention layers use high-frequency features for modular arithmetic. 

Baeumel et al. found that LLMs perform arithmetic tasks digit-by-digit, employing specific circuits for units, tens, hundreds etc. despite some models representing 3 digit numbers as a single token (e.g. `347' would be a single token for Llama 3 8B) \cite{baeumel2025modulararithmeticlanguagemodels, grattafiori2024llama3herdmodels}. 

Aligning with Zhou et al.'s work, which suggests numbers are represented trigonometrically, Kantamneni and Tegmark have shown that LLMs represent numbers on a helix \cite{kantamneni_trig}. This helix is causally implicated in addition and relevant for multiplication and division. For addition, they found that an $a+b$ helix is constructed by manipulating the helices for $a$ and $b$. 


To try to better understand arithmetic errors, Štefánik et al. demonstrate that LLMs share universal, sinusoidal representations of numbers across model families \cite{universalrepresentation2026}. Within this framework they find that the accuracy of arithmetic can be predicted by the sinusoidal fidelity of the output representation. Focusing specifically on addition, Wen et al. defined the `noisy quantisation model' for interpreting errors \cite{shapeofaddition}. In this model errors in large (e.g. ten-digit) addition tend to be `off-by-one' as a result of noise pushing a latent carry signal over a digit boundary. 
Since multiplication can be seen as a series of repeated additions we investigate if a similar noise-driven carry error explains errors in multiplication. 



\section{Experiments}

To investigate multiplication in LLMs empirically, we examine a series of small open weight models: Qwen3.5-2B, Llama-3.2-3B, Gemma-4-E2B, Qwen3.5-0.8B and Qwen2.5-Math-1.5B \cite{qwen3_5_2b, llama3_2_3b, gemma4_e2b, qwen3_5_0_8b, qwen2_5_math_1_5b}. Our code is available at \url{https://github.com/w-bolton/Close-Enough-An-Investigation-into-Multiplication-Errors-in-LLMs}.

We construct a dataset of 16,991 multiplication problems comprised of numbers with an increasing number of digits ($1\text{-digit}\times1\text{-digit}$, $1\text{-digit}\times2\text{-digit}$, $2\text{-digit}\times2\text{-digit}$, $2\text{-digit}\times3\text{-digit}$, $2\text{-digit}\times4\text{-digit}$, $3\text{-digit}\times3\text{-digit}$, $3\text{-digit}\times4\text{-digit}$). The full breakdown is shown in Table \ref{tab:dataset_description}. For the three smallest tiers ($1\text{-digit}\times1\text{-digit}$, $1\text{-digit}\times2\text{-digit}$, $2\text{-digit}\times2\text{-digit}$) every possible operand pair is included since the number of problems is small enough (81, 810 and 8,100 respectively). However, for the four largest tiers, exhaustive enumeration is intractable, e.g. $2\text{-digit}\times3\text{-digit}$ alone has 81,000 pairs. Instead, we sample uniformly at random without replacement from the full set of pairs, capped at 3,000 pairs for $2\text{-digit}\times3\text{-digit}$ and $3\text{-digit}\times3\text{-digit}$, and 1,000 pairs for $2\text{-digit}\times4\text{-digit}$ and $3\text{-digit}\times4\text{-digit}$.
\begin{table}[h!]
\caption{Multiplication dataset by operand-size tier. Three smallest tiers are exhaustive over all possible operand pairs while the remaining four are uniform samples (without replacement) from the full set of pairs.}
\label{tab:dataset_description}
\centering
\begin{tabular}{lrl}
\toprule
Digit size & Count & Sampling \\
\midrule
$1\times1$ & 81    & Exhaustive \\
$1\times2$ & 810   & Exhaustive \\
$2\times2$ & 8{,}100  & Exhaustive \\
$2\times3$ & 3{,}000  & Random sample (of 81{,}000) \\
$2\times4$ & 1{,}000  & Random sample (of 810{,}000) \\
$3\times3$ & 3{,}000  & Random sample (of 810{,}000) \\
$3\times4$ & 1{,}000  & Random sample (of 8{,}100{,}000) \\
\midrule
\textbf{Total} & \textbf{16{,}991} & \\
\bottomrule
\end{tabular}
\end{table}

We use a minimal prompt format (e.g. `$x_1 \times x_2 = $') requiring the models to fill in their answer. All models were run using greedy decoding with a maximum of 10 new tokens. We use regular expressions to extract answers from the models' output, accounting for some models which produce numbers with comma separators. We determine model accuracy by calculating the proportion of correct answers, those where each digit in the answer is correct. We also calculate errors on a per digit basis. We do this since previous work has found that LLMs conduct arithmetic on a per digit basis \cite{baeumel2025modulararithmeticlanguagemodels}.

\begin{figure}[h]
  \centering
  \includegraphics[width=0.75\linewidth]{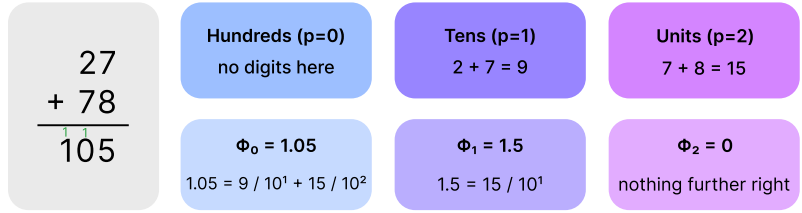}
  \caption{Carry Potential demonstrated for addition}\label{fig:carry_potential_add}
\end{figure}

We borrow the definition of carry potential $\Phi_p$ from Wen et al. (2026) \cite{shapeofaddition},
\begin{equation}
\Phi_p = \sum_{j=1}^{P-1-p}\frac{r_{p+j}}{10^j},
\end{equation}
where $r_{p+j}$ is the sum of the digits at relative position $j$ to the right of the current position $p$ and $P$ is the total number of digit positions in the number. Carry potential represents the `thrust' to carry from previous columns \cite{shapeofaddition}.
Figure \ref{fig:carry_potential_add} shows an example of the carry potential being calculated for a particular addition problem.

This carry potential is then floored to produce a discrete carry used in computation. Wen et al. propose their `Noisy Quantisation' model whereby when carry potential falls close to a digit boundary, neural noise corrupts that carry potential which is then floored to calculate the discrete carry, leading to off-by-one ($\pm1$) digit-wise errors \cite{shapeofaddition}. This hypothesis is supported via their empirical testing, which shows that errors accumulate around the integer boundaries of carry potential and are low elsewhere. Given that multiplication can be viewed as a series of additions, perhaps a similar phenomenon occurs during multiplication.

\section{Results}

\begin{figure}[h]
  \centering
  \includegraphics[width=0.75\linewidth]{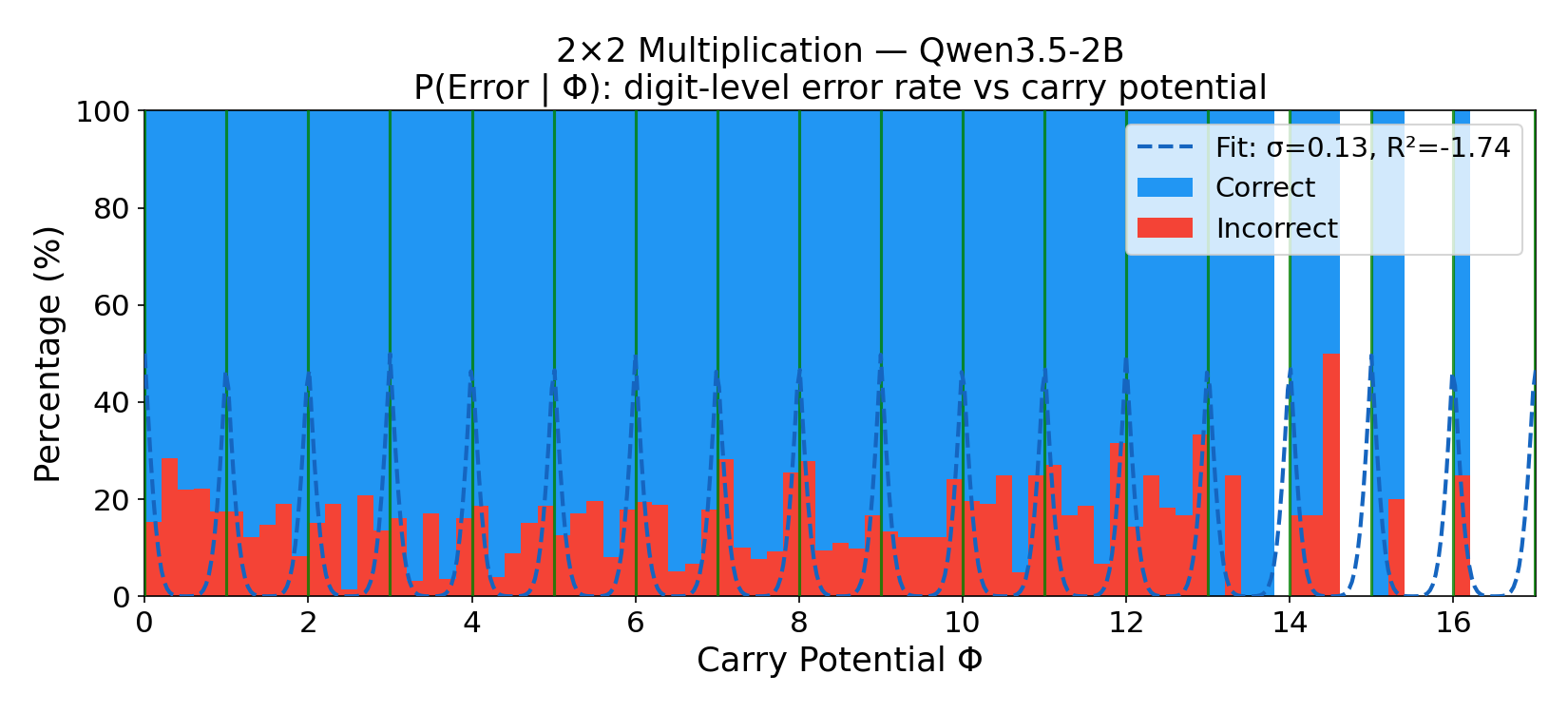}
  \caption{Carry Potential against error percentage for 2-digit$\times$2-digit multiplication using Qwen3.5-2B. The dashed line shows the best-fit noisy quantisation model from Wen et al. (2026), with fitted noise standard deviation of $\sigma=0.13$ and goodness-of-fit $R^2=-1.74$. The negative $R^2$ indicates that the noisy quantisation fit is worse than a flat baseline, confirming the model does not cleanly transfer from addition to multiplication.}\label{fig:carry_potential_vs_error}
\end{figure}

Figure \ref{fig:carry_potential_vs_error} shows the digit-level error rate against carry potential for each 2-digit$\times$2-digit multiplication in our dataset. For multiplication on our dataset we were unable to find the bathtub shape Wen et al. (2026) found for addition whereby errors accumulate when carry potential nears an integer boundary and are very low otherwise \cite{shapeofaddition}. 
This is likely because of how entangled these errors are in multiplication of larger numbers where a series of partial products must be computed and then summed to produce the final result. 

\begin{figure}[h]
  \centering
  \includegraphics[width=0.75\linewidth]{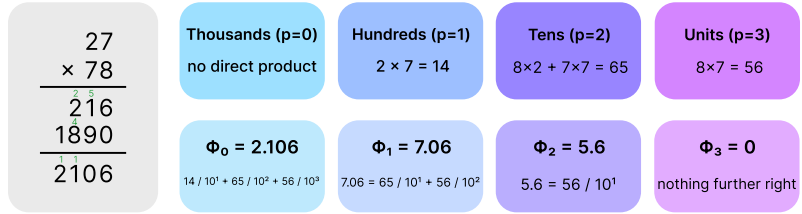}
  \caption{Carry Potential demonstrated for multiplication}\label{fig:carry_potential_mult}
\end{figure}

Another difference between our task of multiplication and theirs of 10-digit addition is that the magnitude of carry potentials is far more variable. Figure \ref{fig:carry_potential_mult} shows how carry potential is calculated for an example multiplication problem; contrasting this with Figure \ref{fig:carry_potential_add} which shows the same calculation for addition, makes clear that the carry potential for multiplication is a much larger and more volatile quantity since it compounds across partial products. The carry potential along any column for addition of 3 numbers is capped at 3 since it can only be as large as the number of digits summed. However, for multiplication of two 2-digit numbers it can be as large as 17. For example in $99\times99$ the tens column raw sum is $(9\times9)+(9\times9)=162$ which combines with the carry from the previous column of $8$ (since $9\times9=81$) to give a carry potential of $170/10=17$. It is possible that this is impeding the reproduction of a clean bathtub shape that Wen et al. found for 10-digit addition \cite{shapeofaddition}.

\begin{figure}[h]
  \centering
  \includegraphics[width=0.55\linewidth]{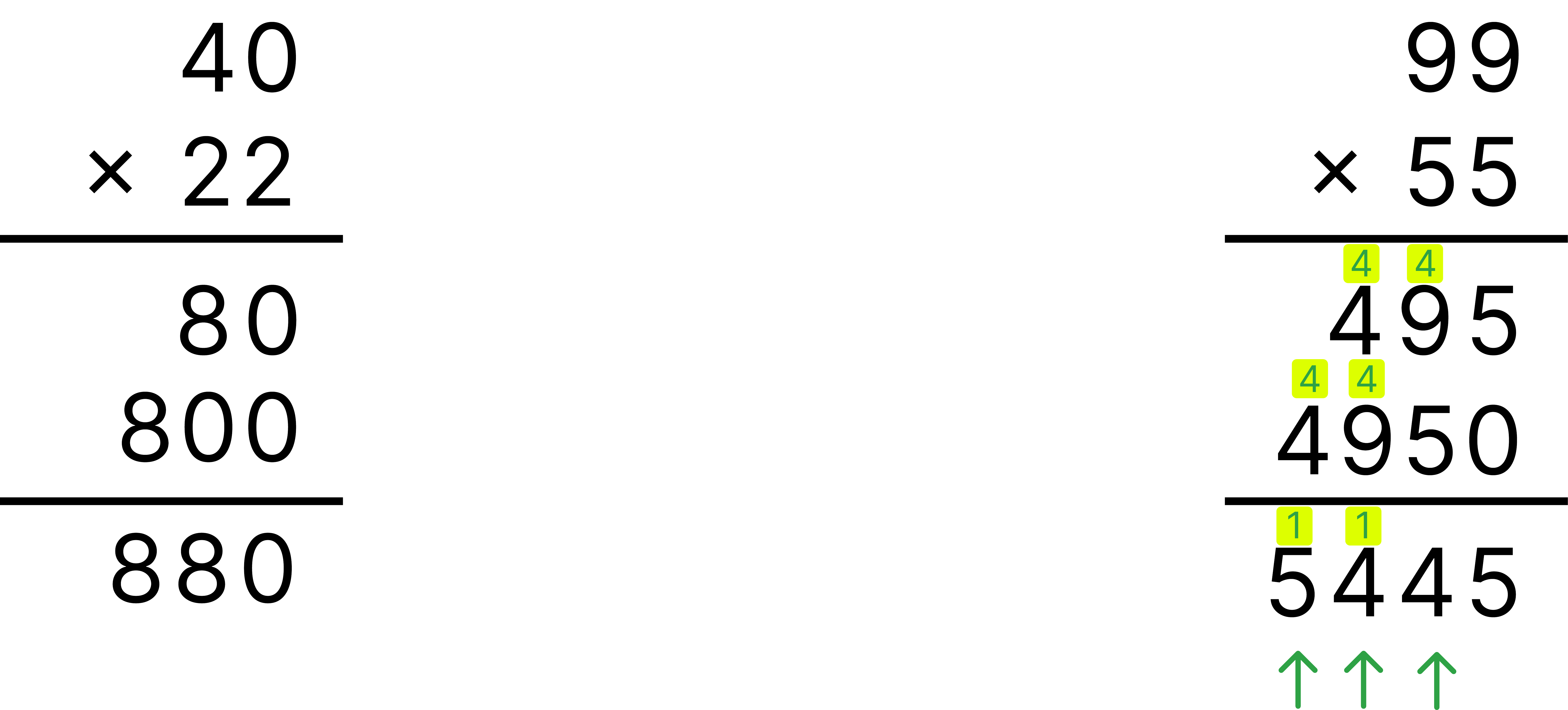}
  \caption{Two examples of long multiplication to show an example of no carry columns (left) and one with three carry columns (right). Carries which contribute to those three carry columns are highlighted in yellow. The green arrows indicate those columns with carries.}
  \label{fig:long_mult}
\end{figure}

\begin{figure}[h]
  \centering
  \includegraphics[width=0.85\linewidth]{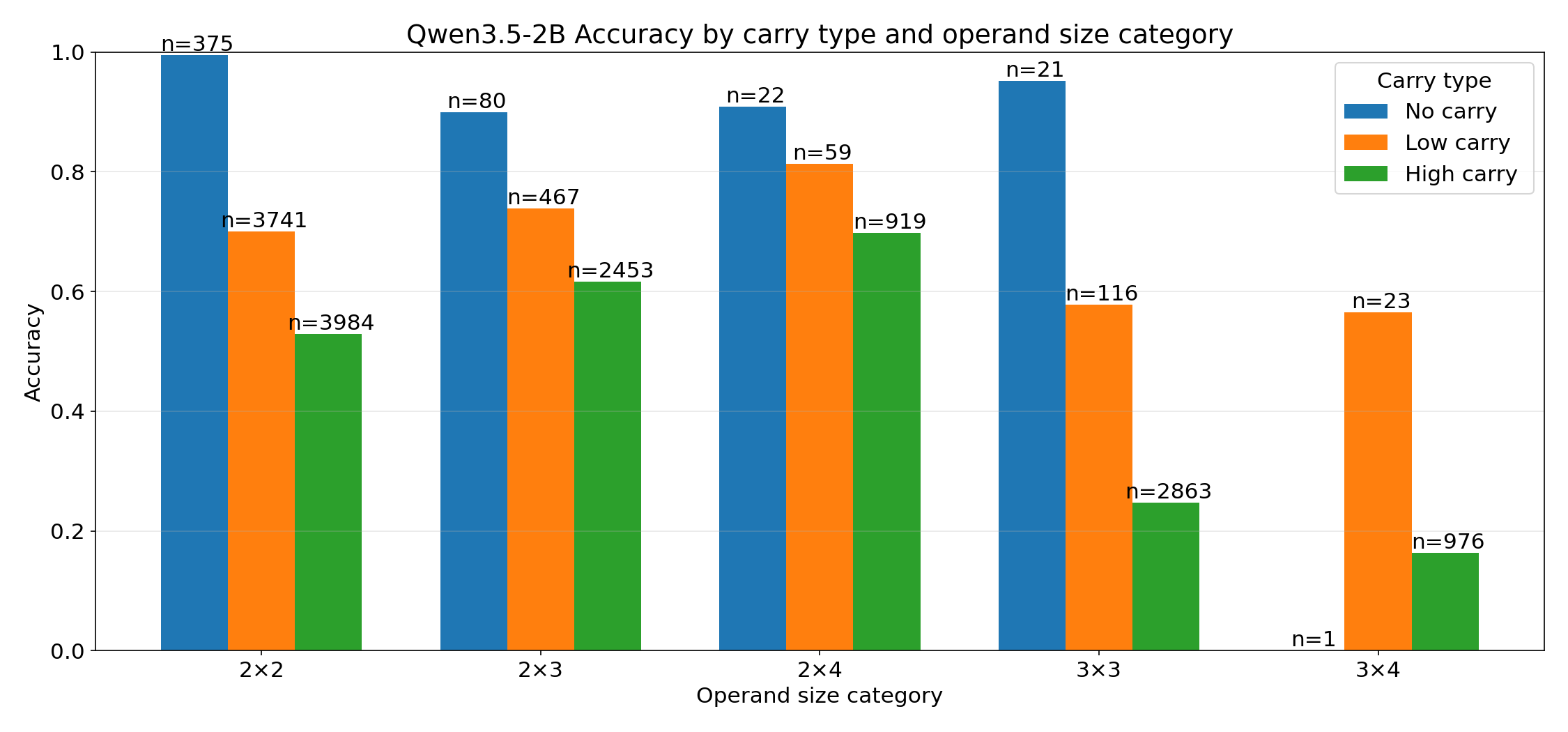}
  \caption{Comparison of the accuracy of multiplications for different operand sizes (i.e. 2-digit x 2-digit up to 3-digit$\times$4-digit) for different carry types (no, low and high).}
  \label{fig:Accuracy_vs_carry}
\end{figure}

Despite this, carry does play a role in the accuracy of multiplication as those calculations without any carries did perform better than those with low ($\le 2$ carry columns) and high carry ($>2$ carry columns). A carry column here is simply a column in the long multiplication which contributes a carry to the next column. For example $40 \times 22=880$ has no carry columns, while $99 \times 55=5445$ has three carry columns so would be classed as an example of high-carry. This is demonstrated in Figure \ref{fig:long_mult} as the first example (left) has no carries whilst the second (right) has a carry in three columns (indicated with green arrows and yellow highlighting). Figure \ref{fig:Accuracy_vs_carry} shows that other than for 3-digit $\times$ 4-digit multiplications where there are too few no-carry scenarios to be informative, accuracy decreases monotonically as the amount of carry increases. This phenomenon is unsurprising given the increased complexity that high-carry scenarios possess. We define the relative error of an incorrect answer as $|\hat{y}-y|/y$ where $\hat{y}$ is the model's prediction and $y$ is the ground truth. The relative error mean was just $1.1\%$ for Qwen3.5-2B indicating that failures tended to have a correct general magnitude. This is consistent with prior work  \cite{zhou2024pretrainedlargelanguagemodels}, showing that LLM numerical representations contain components associated with both approximate magnitude and precise calculation.


\begin{figure}[h!]
  \centering
  \includegraphics[width=0.85\linewidth]{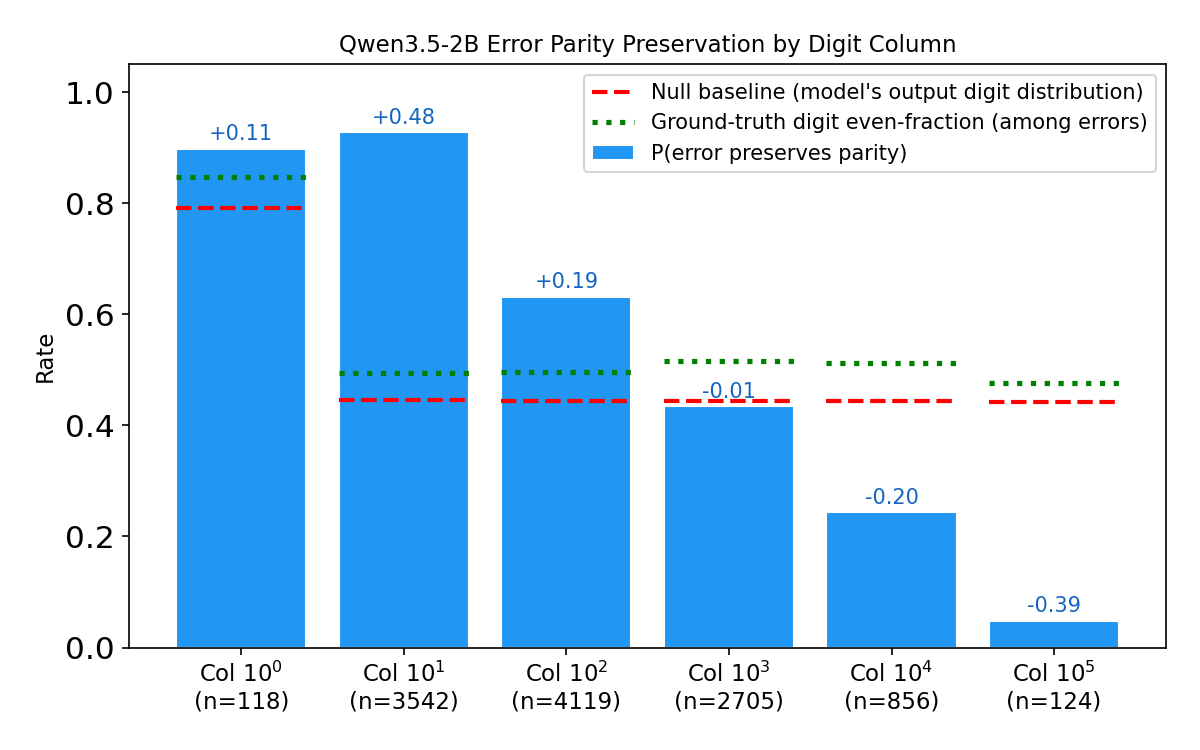}
  \caption{Error parity preservation rate on a per column basis compared to the null baseline and the ground-truth digit even fraction among incorrect digits in that column.}\label{fig:parity_preservation}
\end{figure}

Wen et al. (2026) \cite{shapeofaddition} found for 10 digit addition that errors were often off-by-one. In contrast, we found that for multiplication using small Qwen3.5 models (2B and 0.8B parameter variants) errors in lower order columns, especially the tens-column, were substantially more likely to be of the same parity as the correct answer than not. For Qwen3.5-2B there were 3,542 incorrect digits in the tens-column and these errors had the same parity as the correct digit 92.7\% of the time versus the data-derived chance baseline of 44.4\%. Equivalently, errors in that column were 12.7 times more likely to fall an even distance away from the correct digit than an odd distance. Figure \ref{fig:parity_preservation} shows this parity preservation rate varies across all columns in the dataset. 

If the model were randomly guessing a digit, errors would be parity preserving 4/9 (44.4\%) of the time. Because models can have digit biases we also compute a null baseline from the model's own output digit distribution per column: for an error with the correct answer being $y$, the probability of a parity match is the sum of the probability of all digits with the same parity as $y$ excluding $y$ itself, divided by the summed probability of all digits that are not $y$. We average this over the errors in the column. This gives 44.3-44.5\% for all columns other than the units (79.1\%), where outputs skew even for Qwen3.5-2B. This null baseline is indicated with a red-dashed line in Figure \ref{fig:parity_preservation}. We use a binomial hypothesis test treating each error as a Bernoulli trial with a binary outcome (parity is preserved or not). This assumes independence which does not strictly hold since LLMs generate digits in an autoregressive manner, so each depends on those before it. The p-values should be read as indicative rather than exact. Setting the null hypothesis to be the model based null baseline for each column (errors preserve parity with the probability given by that column's baseline) and the alternative hypothesis that parity preservation is different from that baseline, we reject the null hypothesis for all columns at the 1\% significance level shown in Figure \ref{fig:parity_preservation}, other than the thousands column ($10^3$). The units baseline is high (79.1\%) because the model's outputs skew even, mirroring the ground truth even digit percentage which is 75\% across all products (84.7\% among incorrect units digits). This arises because a product's units digit is only odd when both operands in the units-column are odd ($\text{odd}\times\text{odd}=\text{odd}$); in the other three of the four combinations ($\text{even}\times\text{odd}$, $\text{odd}\times\text{even}$, $\text{even}\times\text{even}$) the product is even. The model might be producing even digits with a higher probability as it is aware of this imbalance from its training data. Accounting for this, the parity preservation rate in the units column (89.8\%) is above the baseline but only modestly. However, in the tens column ground truth even digit percentage is just 54\% across all products (49.5\% among incorrect tens digits) while the parity preservation rate is 92.7\%. The very high order columns ($10^4,10^5$) have drastically few errors which preserve parity, but these are less well statistically powered with only the largest multiplications populating them. Rather than parity preserving errors, for those columns ($10^4,10^5$), 64\% and 85\% of errors respectively are off by one ($\pm 1$).

\subsection{Other Models}
As well as Qwen3.5 0.8B and 2B models we also tested Qwen2.5-Math-1.5B (specifically post-trained for mathematics), Gemma 4 E2B from Google and Llama3.2 3B from Meta \cite{qwen3_5_2b, qwen3_5_0_8b, qwen2_5_math_1_5b, gemma4_e2b, llama3_2_3b}. Although the Qwen3.5-0.8B model showed a similar parity preservation effect in the tens column, it was weaker (69\%) and models from other model families essentially sat at that null, chance baseline. Llama3 represents whole numbers up to 3 digits as a single token whilst Qwen models tokenise on a digit-by-digit basis meaning that the way in which these two models handle multiplication is likely very different. Gemma does tokenise on a digit-by-digit basis but still does not show the same pattern. Perhaps this is due to accuracy being too poor (40\% 2x2, 18\% 2x3, 11.6\% 2x4, 2.6\% 3x3, 0.8\% 3x4) meaning that it is almost guessing at random for larger multiplications. However, Qwen3.5-2B shows the effect for both easier and harder multiplications (when accuracy is high and relatively lower). Parity preservation rate sits at 94\% in the tens-column for simple calculations (1-digit$\times$1-digit, 1-digit$\times$2-digit, 2-digit$\times$2-digit, wrong digits n=2920) when the accuracy is 66.6\%, but is still 91\% in the hundreds-column for 3-digit$\times$3-digit (n=1670) when accuracy is just 26.5\%. We also test Llama3 by forcing it to represent each digit as a single token by adding a space separator between them. Even in this configuration the effect cannot be replicated but again this could be because accuracy is simply too low (25.0\% across all multiplications). Qwen2.5-Math-1.5B shows a mild effect in tens column (67.9\%). Accuracy and parity preservation in the tens-column are shown in Table \ref{tab:model_comparison}.
\begin{table}[h]
\centering
\caption{Multiplication accuracy across similarly sized models tested on the same dataset (n=16,991 and null parity preservation baseline is 44.5-45.2\%)}
\label{tab:model_comparison}
\begin{tabular}{lrrrr}
\toprule
Model & Accuracy (\%) & Tens Parity Preservation\ (\%) \\
\midrule
Qwen3.5-2B & 56.4 & 92.7  \\
Qwen3.5-0.8B & 60.8 & 69.5  \\
Qwen2.5-Math-1.5B & 50.4 & 67.9  \\
Llama-3.2-3B & 25.0 & 45.2 \\
gemma-4-E2B & 28.8 & 44.0 \\
\bottomrule
\end{tabular}
\end{table}

Although Qwen3.5-0.8B achieves higher overall accuracy on our multiplication benchmark than Qwen3.5-2B (60.8\% vs. 56.4\%), Qwen3.5-2B exhibits a substantially stronger parity-preserving error pattern in the tens column (92.7\% vs. 69.5\%). This suggests that overall arithmetic accuracy and the structure of arithmetic errors are not necessarily correlated. The difference may reflect differences in the internal representations or computational strategies adopted by the two models, with the larger model potentially encoding coarse numerical properties such as parity more strongly while not necessarily translating this representation into more accurate digit-level multiplication. Other factors, including differences in training dynamics, post-training, or the allocation of model capacity, may also contribute to the observed discrepancy.

\section{Discussion and Future Work}

\subsection{Discussion}

Our results reveal that multiplication errors in small language models do not manifest as uniform noise. Instead they diverge significantly from existing addition-based error frameworks \cite{shapeofaddition}. While Wen et al. \cite{shapeofaddition} demonstrated that addition errors follow a noisy quantisation model localised tightly around carry boundaries, this phenomenon does not directly transfer to multiplication. In our experiments, errors remain distributed across carry potentials, likely due to the complex entanglement of generating and summing multiple partial products.

Instead, we identify two distinct error regimes across digit order:
\begin{itemize}
    \item \textbf{Low-Order Parity Preservation:} In lower-order columns (most prominently the tens-column, with 92.7\% parity preservation in Qwen3.5-2B versus a 44.4\% null baseline), errors consistently retain the parity of the ground truth. This suggests that parity ($\bmod\ 2$) information is encoded and preserved more robustly than fine-grained digit identity ($\bmod\ 10$), aligning with findings that transformers represent numerical values via periodic Fourier components \cite{zhou2024pretrainedlargelanguagemodels} and helical trajectories \cite{kantamneni_trig}.
    \item \textbf{High-Order Carry Degradation:} In higher-order columns ($10^4, 10^5$), parity preservation drops sharply, replaced instead by off-by-one errors ($>\!64\%$), suggesting that accumulated carry noise dominates the final output tokens.
\end{itemize}

Notably, this strong parity-preservation effect was pronounced in the Qwen family but absent in similarly sized models from Llama and Gemma suites (Table~\ref{tab:model_comparison}). While differences in tokenization granularity (sub-word vs. digit-level) and baseline task accuracy play a role, this discrepancy also suggests that architectural choices or pre-training dynamics in Qwen may prioritize modular periodic numerical circuits more heavily.

\subsection{Future Work}
To build on these empirical observations, future research should explore three primary avenues:
\begin{enumerate}
    \item \textbf{Causal Circuit Analysis:} Employ activation patching and targeted ablation across residual streams and attention heads on parity-varying counterfactual pairs. Training linear probes across intermediate layers will help verify whether the $\bmod\ 2$ representation remains linearly decodable while the $\bmod\ 10$ signal degrades.
    \item \textbf{Cross-Architecture Representations:} Investigate why parity-preserving circuits emerge distinctly across tokenization strategies and training regimes, clarifying why Qwen exhibits higher numerical fidelity than its peers.
    \item \textbf{Unsupervised Activation Steering:} Building on the work of Štefánik et al. \cite{universalrepresentation2026}, who steered output representations toward higher sinusoidal fidelity using known ground-truth targets, future work should test whether answer-agnostic steering objectives can autonomously restore sinusoidal coherence and mitigate arithmetic failure modes at inference time.
\end{enumerate}

\section*{Acknowledgments}

WB is grateful for the EPSRC Vacation Internship funding that supported this work. BT thanks UK Research and Innovation (UKRI) [Grant ref EP/Y030796/1] and the University of Bristol. HD thanks the EPSRC funded VIVO Hub for Enhanced Independent Living (UKRI142) for their support. 

\bibliographystyle{unsrtnat}
\bibliography{ref}

\end{document}